\documentclass[conference]{IEEEtran}
\IEEEoverridecommandlockouts
\usepackage{graphicx}
\usepackage{prettyref}
\usepackage{tabularx}

\newrefformat{sec}{Section~\ref{#1}}
\newrefformat{fig}{Fig.~\ref{#1}}
\newrefformat{table}{Table~\ref{#1}}
\usepackage{cite}
\usepackage{amsmath,amssymb,amsfonts}
\usepackage{algorithmic}
\usepackage{graphicx}
\usepackage{textcomp}
\usepackage{xcolor}
\usepackage{cuted}
\usepackage{capt-of}
\usepackage{booktabs}
\def\BibTeX{{\rm B\kern-.05em{\sc i\kern-.025em b}\kern-.08em
    T\kern-.1667em\lower.7ex\hbox{E}\kern-.125emX}}

\usepackage{fancyhdr}
\begin{document}

\title{Temporal Posed and Spontaneous Gesture Recognition from Electromyography in the Rock-Paper-Scissors Game
\thanks{}
}

\author{
\IEEEauthorblockN{Xin Wei}
\IEEEauthorblockA{
    \parbox{4.2cm}{\centering
    \textit{Nara Institute of Science and Technology} \\
    Nara, Japan \\
    wei.xin.wy0@is.naist.jp
    }
}
\and
\IEEEauthorblockN{Huakun Liu}
\IEEEauthorblockA{
    \parbox{4.2cm}{\centering
    \textit{Nara Institute of Science and Technology} \\
    Nara, Japan \\
    liu.huakun.li0@is.naist.jp
    }
}
\and
\IEEEauthorblockN{Felix Dollack}
\IEEEauthorblockA{
    \parbox{4.2cm}{\centering
    \textit{Nara Institute of Science and Technology} \\
    Nara, Japan \\
    felix.d@is.naist.jp
    }
}
\and
\IEEEauthorblockN{Monica Perusqu\'ia-Hern\'andez}
\IEEEauthorblockA{
    \parbox{4.2cm}{\centering
    \textit{Nara Institute of Science and Technology} \\
    Nara, Japan \\
    perusquia@ieee.org
    }
}
}

\maketitle
\thispagestyle{fancy}

\begin{abstract}
The importance of gesture recognition has been acknowledged in many domains requiring real-time recognition systems. Two requirements for these are fast recognition in multiuser contexts. Therefore, we explored the temporal characteristics of electromyography (EMG) and its accuracy in recognizing gestures in a Rock-Paper-Scissors (RPS) game. Twenty-four participants played RPS in dyads, while a two-channel EMG was recorded from the forearm. We found out that EMG onsets could be detected at least 800~ms before the gesture's visible onset, and that the EMG peaks around 342~ms before the visible onset of the gesture. 
Furthermore, we evaluated self-gesture recognition in both posed and spontaneous gesture conditions. The mean accuracy for posed gestures reached 63.4\%. The model trained on posed gestures achieved 53.6\% for spontaneous gestures, with considerable variation across individuals.
We also checked whether detecting a player's gesture from the opponent's EMG was possible. The peak mean accuracy was 65\%, peaking at 2082~ms after the visual onset of the gesture. This suggests that the opponent's reaction to an observed gesture contains information about the observed gesture due to the dynamics of the interactions while playing. 
The temporal predictive advantage of EMG signals, where muscle activation precedes observable movement, offers potential benefits for applications requiring rapid intent recognition, such as human-computer interaction and assistive technologies. Future work should focus on refining onset detection and reducing the impact of spontaneous movement variability across conditions to improve recognition performance in dynamic and real-world environments.
\end{abstract}


\section{Introduction}
Gesture recognition supports applications such as intuitive user interfaces~\cite{songContinuousBodyHand2012,changExplorationHumanComputer2023}, behavioral analyses of mental states~\cite{lopes-rochaGesticulationIndividualsRisk2023,chenAnalyzeSpontaneousGestures2019}, and sign language recognition~\cite{hanFusionbasedSpatiotemporalConvolutions2021}. The game of Rock-Paper-Scissors (RPS) is a popular choice to benchmark gesture recognition in different settings. For example, in 2002, an RPS module was proposed as a situated module for collaborative gaming between robots and humans~\cite{kandaConstructiveApproachDeveloping2002}.

Computer vision remains the most common approach for detecting and classifying gestures. In the particular case of interactive games such as RPS, an important requirement is to have high-speed sensing that can decide the gesture class in the first milliseconds of the task~\cite{itoTrackingRecognitionHuman2016}. Therefore, other sensing modalities are gaining attention. 
Leap Motion has been used at a sampling rate of 120~Hz to detect posed RPS gestures with up to 93\% accuracy~\cite{brockDevelopingLightweightRockPaperScissors2020}.
Electromyography (EMG) is another method often used for gesture detection. EMG can be sampled at higher frequencies than regular cameras. Its ability to detect muscular activity before a gesture becomes visible is promising for earlier and more accurate gesture onset detection~\cite{perusquia2021fg}. However, the exact relationship between EMG signals, visual onset, and recognition accuracy requires further exploration, particularly in terms of identifying the earliest point at which classification surpasses random chance.

Furthermore, gestures are social expressions, and the importance of focusing on the relationship between the gesture producers has been highlighted in the past~\cite{alghowinem2021}. However, most existing research has focused on gestures as performed by individuals, often neglecting the social dimension in which those gestures may occur. Understanding how gestures manifest within dyadic interactions —such as in collaborative tasks or games— could offer novel insights and expand the current scope of gesture recognition methodologies. 

Therefore, we present an analysis of how the actions of one person can be detected from the behavior of another person interacting with them. This has already been attempted in the frame of emotion recognition. Previous work has proven that it is possible to classify the type of smile a person looks at from the observer's skin conductance using temporal features~\cite{hossainUsingTemporalFeatures2020}.
The contributions of this work are:
\begin{itemize}
    \item A direct evaluation of recognition performance for three posed and spontaneous gestures.
    \item A report of the temporal difference between EMG and visible onset of the gestures for the \textit{Extensor Carpi Radialis Longus} and the \textit{Flexor Carpi Ulnaris} arm muscles.
    \item A detailed temporal analysis of the EMG-based gesture recognition with respect to the visible onset.
    \item An analysis of the opponent's gesture from the observer's EMG.
    \item An open dataset for future investigations.~\footnote{https://doi.org/10.17605/OSF.IO/FMREA}
\end{itemize}

\section{Related Works}
Gesture recognition implementations of the game RPS are a popular tool in social robotics for collaborative gaming.
Previously, hand gesture recognition systems using images achieved a recognition rate of 92\% for 35 gestures, but after an average elapsed time of 2.76~s~\cite{panwarHandGestureRecognition2011}.
Ito et al. (2016) used a high-speed vision system to detect the three gestures in RPS~\cite{itoTrackingRecognitionHuman2016}. They identified that a robot hand response of 300~ms or less and a wide-area camera were necessary. They introduced a projection-based approach that ensured sufficient illumination. The hand tracking was done by detecting first the palm as the center of gravity of the hand, then assessing if the hand was open, and finally identifying the number of extended fingers. The system's performance peaked at 330~ms, and its recognition algorithm needed windows of 180~ms on average to compute the class of a gesture. They reported performance high enough to beat a human opponent with a robot.

\begin{table}[t]
\centering
\caption{EMG-based related works on RPS gesture classification.}
\renewcommand{\arraystretch}{1.2}
\resizebox{\columnwidth}{!}{%
\begin{tabular}{l|c|c|c}
\hline
\toprule
\textbf{Reference} & \textbf{Dataset / Scenario} & \textbf{Gesture Type} & \textbf{Accuracy (\%)} \\
\hline
\midrule
Jang et al., 2012 & 6 participants & Spontaneous (online) & 48.2 \\
Funabashi et al., 2015 & 1 participant & Posed (offline) & 77.1 \\
Funabashi et al., 2015 & 1 participant & Posed (online) & 74.8 \\
Ploengpit et al., 2016 & 1 participant & Posed (offline) & 78 \\
Gang et al., 2017 & 1 participant & Posed (offline) & 97 \\
This work & 24 participants & Posed (offline) & 63.4 \\
This work & 24 participants & Spontaneous (offline) & 53.6 \\
\hline
\bottomrule
\end{tabular}
}
\label{tab:related_comp}
\end{table}

Previous studies have also investigated the feasibility of recognizing or predicting RPS hand gestures using electromyography (EMG) signals. 
Jang et al. (2012) proposed a real-time prediction method for ternary classification of RPS gestures using EMG signals recorded from electrodes placed around four forearm muscles~\cite{jang_rock-paper-scissors_2012}. 
Participants played the game spontaneously against a computer during the experiment.
Their approach utilized the early burst of EMG signals, which occurs approximately 100~milliseconds before the physical movement onset detected using a data glove system, to make predictions.
The exact prediction performance was not reported numerically but illustrated using boxplots.
Based on visual estimation of the medians from each boxplot and averaging them across all three gestures and four muscles, the overall mean accuracy was approximately 48.2~\%.
Funabashi et al. (2015) developed a finger motion recognition system for three rock-paper-scissors gestures and a neutral gesture by implementing both offline and online versions~\cite{funabashi_-line_2015}.
Participants performed the gestures following the given instructions.
The offline implementation, which used neural networks, achieved a mean accuracy of 77.1\% across the four gestures.
In contrast, the online version using Principal Component Analysis (PCA) and a nearest neighbor classifier yielded a mean accuracy of 74.8\%. 
Interestingly, the recognition accuracy for the scissors gesture was relatively low in the online evaluation, with a mean accuracy of only 22.7\%.
Ploengpit et al. (2016) used an armband-type EMG device and a decision tree classifier to recognize three RPS gestures~\cite{ploengpit_rock-paper-scissors_2016}.
EMG data were collected from both arms of a single participant while holding each gesture for 5 to 25~s across 10 rounds.
Holding the pose for 15~s yielded the highest classification performance among the tested durations.
They reported a recognition accuracy of 78\%. However, since the same dataset was used for both training and testing, the reported performance may be overestimated.
Gang et al., (2017) employed a multilayer perceptron to classify three RPS gestures using EMG signals~\cite{gang_classification_2017}.
Their experiment involved performing each gesture successively, with each pose sustained for 2 to 3~s.
 Ten-fold cross-validation achieved 97\% accuracy. 
However, the training and test sets were split by ratio, meaning that data from each gesture appeared in both sets.
This approach may inflate performance and limit generalizability to unseen executions.

We summarized EMG-based related works on RPS gesture classification in~\prettyref{tab:related_comp}.
This review reveals a gap in recognizing spontaneous gestures, which more closely reflect how RPS is played in the real world.
In this study, we explored both posed and spontaneous gesture recognition.

Several works have quantified the difference in timing between EMG activity and visual behavior. Regarding the face's \textit{Zygomaticus Major} muscle, 230~ms~\cite{cohnTimingFacialMotion2004a} and 374~ms~\cite{perusquia2021fg} have been reported as lead time over visual behavior. However, physiological studies also reported muscle specificity with different latencies counting from cortical activation. Therefore, investigation for other muscles is necessary~\cite{naishEffectsActionObservation2014}. 
Huber et al., found that EMG mean onset latencies were between 106~ms and 171~ms. Neck extensors responded with the shortest latency, and in most cases, right-side muscles responded faster in the frame of vehicle driving~\cite{huberMuscleActivationOnset2013}.
In this study, we investigate the latencies of two forearm muscles. 

\section{Methods}
\subsubsection{Participants}
We recruited 24 participants (7 female, mean age 28.2 years, SD = 4.4) for 12 experiments, with two participants playing in the same session.
The experimental procedure was explained beforehand, and we obtained informed consent from all participants.
The study followed the guidelines of the Declaration of Helsinki and was approved by the Institutional Review Board of NAIST, with review number 2024-I-35.

\subsubsection{Experiment design}
We used a \(2\times3\) within-subjects design.
The independent variables were 
hand gesture type (Rock, Paper, Scissors), and player (self, opponent).

\subsubsection{Stimuli}
Sound stimuli were used to instruct the participants on what gestures to perform and when.
Beep sounds marked the start and end of each round, with an additional middle beep used during the free-play stage to signal participants to finish their chosen gesture before this point.
Audio instructions in Chinese, English, and Japanese served as instructions and prompts for playing RPS.
The language was matched to the participant's native language.

\subsubsection{Measurements}
EMG data and video recordings were collected during the experiment.
The camera was positioned to capture the participants' hand movements without including their faces.
We utilized ELAN (EUDICO Linguistic Annotator) to extract visible onsets and offsets based on the video’s relative timestamps.
The annotation process involved three annotators labeling each trial's first visible movement around the start beep as the onset and the first frame of three consecutive motionless frames after the participant began returning to the relaxed gesture as the offset.
Each annotator labeled different experiment sessions. The annotation method was discussed among annotators to ensure agreement. 
The EMG was recorded using two channels per participant from their right hand. The first is on the muscle \textit{Extensor Carpi Radialis Longus}, and the second is on the \textit{Flexor Carpi Ulnaris}. Both of these muscles aid the extension and flexion of the hand. The reference was placed on the heaf of the \textit{Ulna} bone, Olecranon, or elbow.
The experimenter labeled the participants' gestures during the experiment's free play task.

\begin{figure}[!t]
    \centering
    \includegraphics[width=\columnwidth]{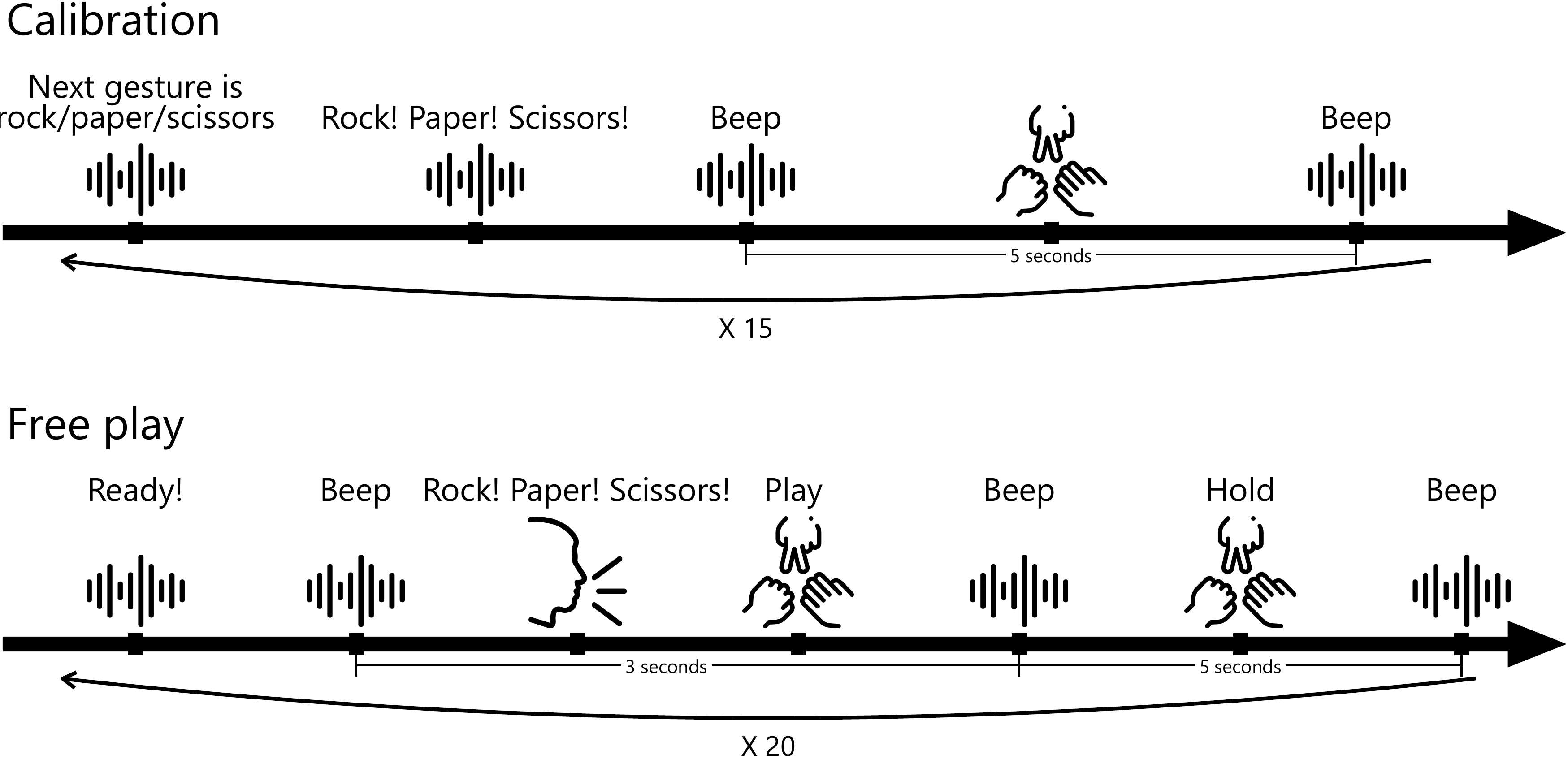}
    \caption{Tasks overview. In the calibration task, both participants performed the same gesture as instructed by sound stimuli. The order of the gestures was randomized, and each category was balanced. In the free play task, the participants played spontaneously, choosing what gesture to perform, but the timing of the gesture display was instructed with sound cues.}
    \label{fig:task}
\end{figure}

\subsection{Task}
The experiment was divided into two tasks: calibration and free play. An overview of the tasks is shown in \prettyref{fig:task}.

\subsubsection{Posed Gesture Calibration}
The goal was to collect a balanced sample of the three RPS gestures. 
Every gesture was repeated five times, resulting in a total of 15 trials, presented in a randomized order. 
Each trial began with an audio instruction specifying the gesture both participants should perform, followed by a voice prompt saying, “Rock! Paper! Scissors!”
A beep then signaled participants to execute the instructed gesture immediately.
Each gesture was held for five seconds until another beep indicated the end of execution.

\subsubsection{Spontanueous Gesture Free Play}
This task aimed to gather spontaneous gestures without instruction.
The participants freely played 20 rounds of RPS with each other.
Each trial began with an audio cue, “Ready!” followed by a beep signaling the start of the game.
Participants cued “Rock! Paper! Scissors!” themselves to mimic a real gameplay experience.
They were required to display their gestures before the second beep, which occurred three seconds after the first.
After the second beep, they held their poses for five seconds until the third beep marked the end of the trial.
The experimenter recorded both participants’ gestures after the third beep before proceeding to the subsequent trial.

Participants completed a practice session to familiarize themselves with the tasks, which included two repetitions of each gesture (totaling six trials) in the calibration task and five rounds of gameplay in the free play task.
Participants were instructed to avoid shaking their hands while playing the game to ensure signal quality.
We also required participants to use a slightly open palm as their resting gesture to be able to record the change from the resting state baseline to each gesture peak, as well as the offset going from the peak gesture to the resting baseline.

\begin{figure}[!t]
    \centering
    \includegraphics[width=\columnwidth]{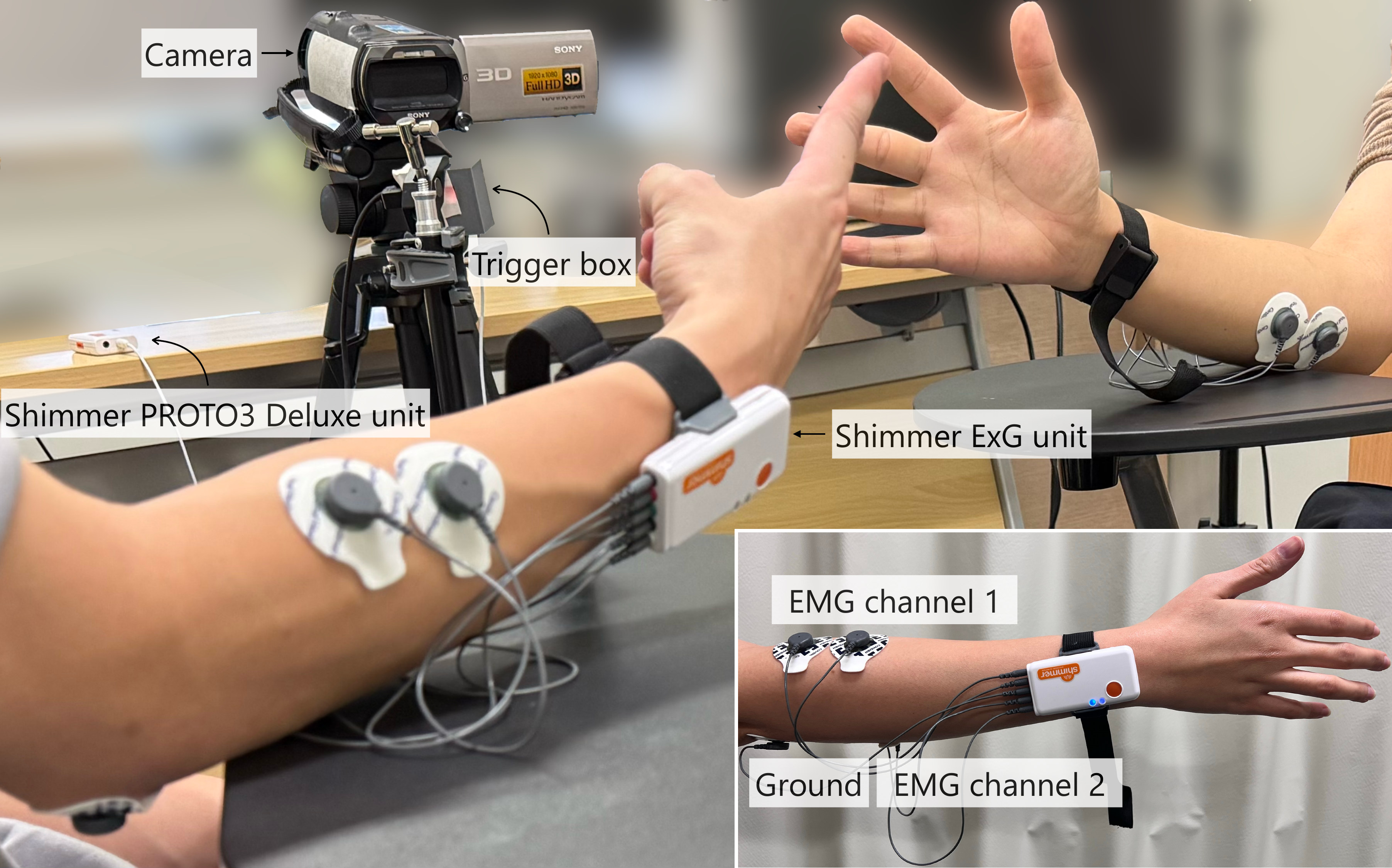}
    \caption{Experimental setup. Two participants sat facing each other with their arms resting on the chair’s surface. A camera recorded their hands, excluding faces. ExG units were worn on the wrist, with electrodes placed around target muscles and a reference point. A trigger box containing an Arduino Nano with an LED light was connected to a Shimmer PROTO3 Deluxe unit, sending trigger events via serial port to the Shimmer system, while LED flickers marked triggers in the video.}
    \label{fig:app}
\end{figure}

\begin{figure*}[!t]
    \centering
    \includegraphics[width=\textwidth]{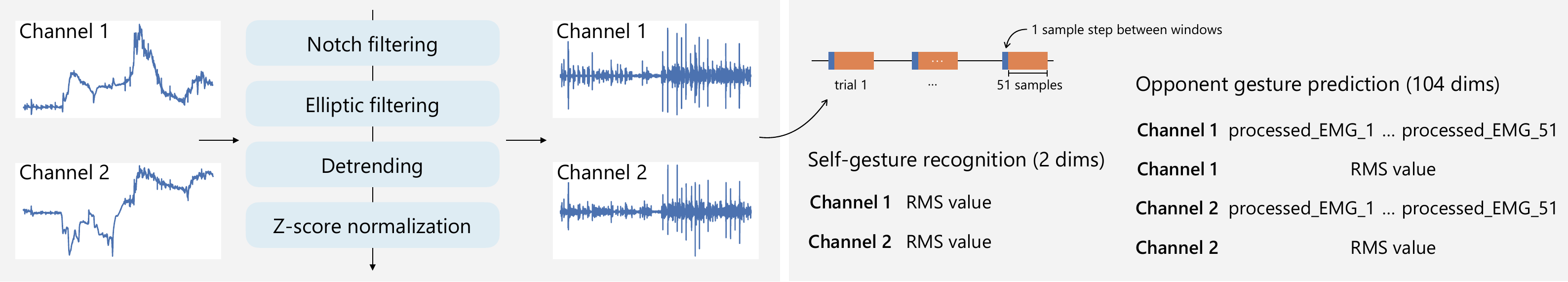}
    \caption{EMG Signal Processing Pipeline and Feature Extraction. Raw EMG signals from two channels underwent preprocessing steps, including notch filtering, elliptic filtering, detrending, and Z-score normalization. The processed signals were then segmented into overlapping windows, each containing 51 samples with a step size of one between adjacent windows. Task-specific features were extracted from two EMG channels: two RMS values for self-gesture recognition, and 104-dimensional vectors included processed EMG values and RMS values for opponent gesture prediction.}
    \label{fig:process}
\end{figure*}

\subsection{Apparatus}
The experimental setup and electrode placement are shown in \prettyref{fig:app}.
Two participants sat in front of each other during the two tasks.
They were seated in a tablet-arm chair, resting their arms on the attached surface.

A Sony HDR-TD10 handheld camera (1920×1080 resolution, 30 FPS) was used to record participants’ hand gestures.
We used two Shimmer ExG units to record EMG signals, each capturing two data channels.
A Shimmer PROTO3 Deluxe unit recorded event triggers sent via a serial port connected to an Arduino Nano board.
The Arduino board, equipped with an LED indicator, was placed inside a grey box and fixed with a tripod clamp in front of the camera.
Thus, the camera could capture the LED flickers, allowing synchronization between the video recordings and EMG signals.
The three Shimmer units were programmed with LogAndStream firmware before the practice session so that we could verify cable connections by streaming and visualizing EMG signals via Bluetooth.
Before the main experiment, the three Shimmer units were programmed with SDLog firmware and synchronized with each other using their independent onboard clocks related to the master unit’s clock.
We designated the PROTO3 Deluxe unit as the master unit.
All Shimmer devices recorded data at a sampling rate of 512~Hz.
A PsychoPy program controlled the experiment and generated audio instructions and beeps. This software was run on a MacBook Pro (16-inch, 2021) equipped with an Apple M1 Max chip and 64 GB of memory.

\subsection{Procedure}
The participants were first explained about the experiment's electrode placement, stimuli, measurements, and goals. 
Then, they provided informed consent.
Participants were asked to move their right hand to locate the \textit{Extensor Carpi Radialis Longus} and the \textit{Flexor Carpi Ulnaris} muscles. At the same time, the experimenter palpated the participant's arm to find the muscles moving. 
Electrodes were then placed along the identified muscle after cleaning the skin with alcohol pads.
Next, Shimmer devices were connected, the camera recording was started, and the practice program was launched.
Once the practice session was finished, the experimenter answered any questions from participants before proceeding to the main session.
At the end of the experiment, the participants were rewarded with snacks.

\section{Analysis}

\subsection{Data Exclusion}
One participant missed one trial in the calibration stage and one in the free play stage, while the other participant in the same session missed one trial in the calibration stage and three trials in the free play stage. 
Additionally, one EMG channel failed to record usable signals due to a device problem. 
Data from these trials, as well as from the participant wearing the malfunctioning EMG device, were excluded from all subsequent analyses.

\subsection{Preprocessing}
The preprocessing and feature extraction pipeline for the EMG signals is shown in \prettyref{fig:process}.
We applied notch filters at 60~Hz, 120~Hz, 180~Hz, and 240~Hz to remove power line noise and its harmonics. Each filter removed these frequencies within a \(\pm\)3~Hz range around the target frequency.
This was followed by an elliptic filter with a passband of 5~Hz to 250~Hz.
The upper bound was set based on the Nyquist frequency, given the sampling rate of 512~Hz, while the lower bound was chosen to exclude movement artifacts and low-frequency noise.
We then applied linear detrending to eliminate trends caused by sweat and biosignal changes.
We normalized the training set in each evaluation by calculating Z-scores. Afterward, the computed mean and standard deviation were used to normalize the test set.
The sliding window moved over the normalized data with a window length of 100~ms (51 samples) and a step size of one sample~\cite{perusquia2019}.
Features were extracted from the two EMG channels using two distinct strategies, depending on the task.
For the self-gesture recognition task, using detailed signal inputs could lead to overfitting, where the classifier performed well on training data but failed to generalize to unseen data.
To mitigate this, each feature vector consisted of two RMS values, one from each channel, calculated from each window.
In contrast, the goal of the opponent gesture prediction task was to infer the opponent’s gesture based on the observer’s EMG signals.
Considering these signals do not directly reflect the opponent’s muscle activity, overfitting was less of a concern. 
Therefore, each feature vector was constructed to include 51 normalized EMG samples per channel along with their RMS values, resulting in a 104-dimensional feature vector.

\begin{figure}
    \centering
    \includegraphics[width=\linewidth]{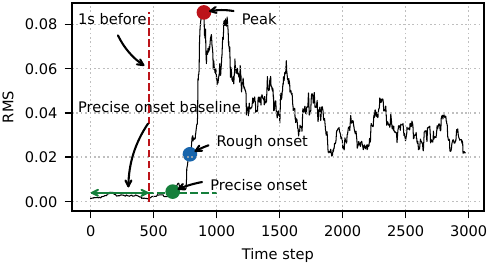}
    \caption{Illustration of the EMG onset detection process using RMS values. The black curve represents the RMS values computed from a trial in the calibration task and the one-second pre-trial segment. The peak point (red) corresponds to the maximum RMS value within the trial. The rough onset (blue) is identified as the first sample exceeding 25\% of the peak RMS value. The precise onset baseline (green) is derived from the one-second pre-trial segment, calculated as the mean RMS value plus two standard deviations. The precise onset is determined as the sample occurring immediately after the closest sample that falls below this baseline.}
    \label{fig:onset-detection}
\end{figure}

\subsection{EMG Onset detection}
The EMG onset detection method is illustrated in \prettyref{fig:onset-detection}.
We detected EMG onsets in each trial from two EMG channels using a two-step approach.
The method considers two-time ranges: (1) the trial itself and (2) the one-second pre-trial interval.
We created sliding windows in both ranges using the same window size and step as described in the preprocessing stage, and RMS values were computed for each window.
In the first step, we set a dynamic threshold of 0.25 times the maximum RMS value within the trial to identify the muscle activation period associated with gesture execution.
The first instance in the trial where the RMS exceeded this threshold was considered a rough onset.
However, this onset might be delayed compared to the actual muscle activation.
To refine the onset detection, we used the one-second pre-trial segment as a baseline for relaxation. 
A secondary threshold was defined as the mean RMS plus two times the standard deviation in the baseline segment. 
The final EMG onset was determined as the sample one after the closest RMS value in the trial range that fell below this secondary threshold before the rough onset. 
If no such value was found, the onset was set to the earliest available timestamp in the trial.

\subsection{Model training}
We used a consistent machine learning architecture while implementing different strategies for designing the input of models to test the hypotheses of interest.
For the model architecture, we utilized Autogluon, an open-source AutoML and ensemble learning framework, which includes 14 models: gradient boosting models (LightGBM variants, XGBoost, CatBoost), neural networks (NeuralNetFastAI, NeuralNetTorch), decision tree-based ensembles (RandomForest, ExtraTrees), K-Nearest Neighbors (KNN), and a weighted ensemble model~\cite{erickson_autogluon-tabular_2020}.
To balance model performance and computational efficiency, we set the parameter presets as ``medium\_quality''.

\subsubsection{Self-gesture recognition}
We first examined the baseline hand gesture recognition performance in our dataset.
For this analysis, we employed a leave-one-trial-per-class-out strategy.
In the calibration task, participants performed each of the three gestures five times.
One trial per gesture was randomly selected as test data. The remaining 12 trials were used to train the classification model using data collected between the first and second beeps (i.e., when the participants executed the instructed gestures).
We repeated this procedure five times without reusing any combination of selected test trials.
In addition, we used data from three selected trials to evaluate posed gesture recognition, and all trials from the free-play task to assess spontaneous gesture recognition.
Each five-second trial in the calibration task contained an average of 2512.9 sliding windows (SD = 1.8), while each 8-second trial in the free-play task contained an average of 4048.7 windows (SD = 3.4).
This slight difference arises from minor fluctuations in the effective sampling rate.
In addition to evaluating overall classification accuracy, we analyzed how accuracy evolved over time during spontaneous gesture recognition. Models were trained on data from the 15 calibration-task trials and tested on the 20 free-play trials. This allowed us to assess temporal performance in a more naturalistic setting where participants often performed additional movements like shaking hands with the opponent before displaying an RPS gesture.

\subsubsection{Opponent gesture prediction}
To investigate the possibility of predicting an opponent’s gesture based on the observer's EMG activity, we reversed the labels of the two players within the same session.
This approach assumes that one participant’s EMG signals may contain predictive cues about their opponent’s upcoming gesture. 
Additionally, the participant’s post-event behaviors, influenced by observing the opponent’s gesture or reacting to the outcome of the game, may also provide informative signals.
We employed a leave-one-out cross-validation strategy, using data from each trial in the free play as the test set while the remaining 19 trials served as the training set.
On average, the training data contained 76,329.9 sliding windows (SD = 2,360.9), while the test data had 4,048.7 sliding windows (SD = 3.4).
The large standard deviation in the training data is due to the exclusion of invalid trials.
For the test data, since each test set consists of a single trial, any invalid trial is skipped.

\section{Results}


\begin{figure}
    \centering
    \includegraphics[width=\linewidth]{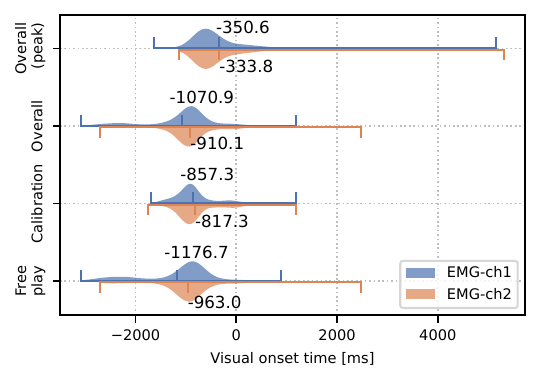}
    \caption{Distribution of EMG onset differences across conditions. Violin plots show the differences between EMG-detected onsets from two channels and visible onsets (blue: EMG onset from channel~1 minus visible onset, orange: EMG onset from channel~2 minus visible onset) across the overall experiment, calibration task, and free play task.}
    \label{fig:onset-comparison}
\end{figure}
\begin{figure*}[!t]
    \centering
    \includegraphics[width=\textwidth]{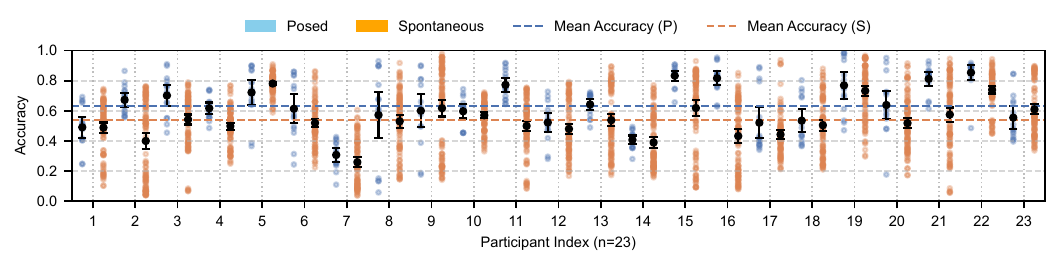}
    \caption{Accuracy distribution of self-gesture recognition across participants in posed and spontaneous conditions. Each dot represents the classifier’s mean accuracy when tested on data from a single trial. The posed condition includes 15 trials per participant (from repeated leave-one-trial-per-class-out evaluation), while the spontaneous condition includes 100 free-play trials (except for participants with invalid trials). Error bars indicate the 95\% confidence intervals of trial-level accuracies for each participant. Dashed lines show the grand mean accuracy for each condition across all participants. The chance level is 33.3\%.}
    \label{fig:individual-performance}
\end{figure*}

\subsubsection{EMG onset and visible movement onset comparison}
To compare the differences between EMG onsets and visible onsets, we aligned both to a unified timestamp format using triggers recorded by the Shimmer device and LED flickers recorded in the videos.
We computed the differences by subtracting the visible onset timestamps from the EMG onset timestamps of two channels. 
The differences across the calibration task, free play task, and the overall experiment are presented in \prettyref{fig:onset-comparison}.
The mean differences in the overall condition (-1070.9~ms for channel~1 and -910.1 ms for channel~2) and all reported mean differences were negative, indicating that EMG activity consistently precedes visually observed movement.
In the calibration task, the mean differences (-857.3 ms, -817.3 ms) were relatively smaller compared to the free play task (-1176.7 ms, -963 ms), suggesting a shorter delay between EMG activation and visible movement in the calibration condition.
In addition, we calculated and presented the RMS peak difference by subtracting the visible onset timestamp from the RMS peak timestamp in each trial across the entire experiment.
The mean differences for the two channels were -350.6 ms and -333.8 ms, respectively.
These results show a temporal sequence: EMG onset occurs first, followed by the moment of maximum muscle activity, and finally, visible movement.




\subsubsection{Self-gesture recognition} 

\begin{figure}
    \centering
    \includegraphics[width=\linewidth]{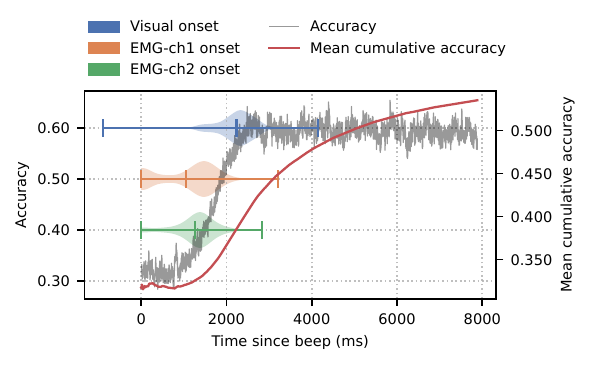}
    \caption{Accuracy trends of cross-task hand gesture recognition relative to onset timings. The plot shows the accuracy (black) and mean cumulative accuracy (red) across participants and trials relative to the first beep. Overlaid distributions represent visual onsets (blue), EMG onsets from channel 1 (orange), and EMG onsets from channel 2 (green).}
    \label{fig:preserve-result}
\end{figure}
\prettyref{fig:individual-performance} presents an overview of self-gesture recognition performance in both posed and spontaneous gesture conditions.
The mean accuracy for posed gestures (63.4\%) is higher than that of spontaneous gestures (53.6\%).
In addition, a high degree of individual differences in performance was observed among participants.
\prettyref{fig:preserve-result} illustrates the temporal evolution of recognition accuracy in the spontaneous hand gesture classification task.
The accuracy used in this and the following results section was calculated as the proportion of correct predictions made at each time step (1.95~ms intervals) across participants and trials.
The mean cumulative accuracy was computed as the average accuracy from the first beep up to each time step
Both accuracies remained relatively low in the early phase, i.e., 0 to 500 ms, when participants were expected to cue “Rock Paper Scissors!”.
However, the performance gradually improves as the game progresses, reaching a peak accuracy of 65.4\% at 3602~ms after the mean visual onset, and a final cumulative accuracy of 53.6\%.
By incorporating onset information, we found recognition accuracy improves immediately following the EMG onsets, indicating muscle activity provides a useful early indicator for gesture recognition.
The visual onset occurs later, aligning with a subsequent stabilization in accuracy.

\begin{figure}
    \centering
    \includegraphics[width=\linewidth]{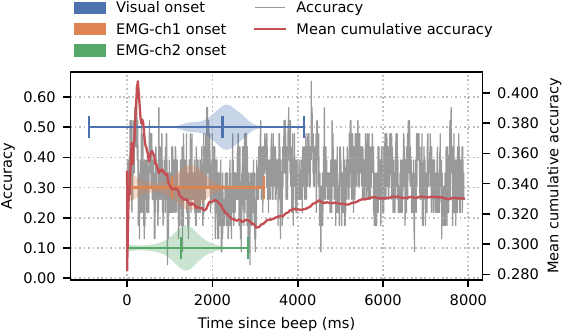}
    \caption{Accuracy trends of opponent gesture prediction relative to onset timings. The plot shows the mean accuracy (black) and mean cumulative accuracy (red) across participants and trials relative to the first beep. Overlaid distributions represent visual onsets (blue), EMG onsets from channel 1 (orange), and EMG onsets from channel 2 (green).}
    \label{fig:reverse-label}
\end{figure}

\subsubsection{Opponent gesture prediction} 
We presented the evolution of prediction accuracy over time when predicting an opponent’s gesture by analyzing the observer’s EMG activity in \prettyref{fig:reverse-label}.
The accuracy initially peaks at 60.9\% at 115~ms after the first beep (0~ms), with cumulative accuracy reaching 40.8\% at 129~ms.
A sharp decline in cumulative accuracy is observed afterward within the first 500~ms.
The decline then stabilizes around the EMG onsets, and the accuracy steadily improves after the visual onset.
The accuracy eventually peaks at 65.2\% at 2082~ms after visual onset.

\section{Discussion}
We investigated the recognition and prediction performance of three hand gestures in the RPS game using EMG signals. We trained gesture recognition models using data from instructed displays and evaluated their performance on both instructed and spontaneous displays.
We explored the feasibility of predicting an opponent’s gesture from the observer’s EMG signals. 
Finally, we analyzed the temporal differences between EMG activation and visible movement of gestures. 

The onset timing difference results reveal a latency between visual onsets and EMG onsets.
Our findings indicate that visual onsets occur approximately 990.5~ms after EMG onsets.
This delay is expected, as muscle activity precedes the actual movement execution, serving as the driving force behind motion.
The muscles targeted in this study are relatively large, and the hand is positioned further from these muscles, potentially explaining the earlier detection of EMG signals compared to similar assessments conducted on facial muscles~\cite{cohnTimingFacialMotion2004a,perusquia2019}.
The delay was larger in the spontaneous part of the game, which might be due to dynamic changes in movement selection during gameplay.

We reported accuracy trends over time with detected EMG onsets and human-labeled visible onsets.
The self-gesture recognition results exhibit a relationship between accuracy improvement and the detected onsets, while the opponent gesture prediction results show a peak accuracy early in the trial.
However, the recognition accuracy obtained in this study is generally lower than that reported in some related works.


\subsubsection{Factors affecting recognition accuracy} 
The lower accuracy observed compared to related works may be attributed to differences in evaluation strategies
We ensured that each trial was kept intact and not split between training and test sets.
This trial-level separation prevents data leakage but may result in greater variability between training and test samples.
Furthermore, the free-play task was designed to mimic real gameplay conditions, allowing us to capture the natural interactions and real reactions between participants.
While this approach provides a more ecologically valid setting, it also introduces greater variability in execution due to its inherently less controlled nature.
Controlled posed gestures lead to higher accuracies, as demonstrated in our study, and the results reported by previous works~\cite{gang_classification_2017,ploengpit_rock-paper-scissors_2016,funabashi_-line_2015}.
In addition, many participants exhibited differences in arm pose between tasks, which could potentially lead to inconsistencies in sensor data.
Moreover, since the calibration task consistently preceded the free-play task, muscle fatigue effects may have contributed to decreased signal reliability over time.

\subsubsection{Accuracy trends over time}
Accuracy in~\prettyref{fig:preserve-result} demonstrates a gradual improvement leading up to the visual offset, followed by a decline in performance.
The initial increase in accuracy is associated with the peak of EMG activity, where signal amplitude is highest, and gesture-related features are most distinguishable.
As muscle activation reaches its peak, the classifier benefits from clearer and more separable signal patterns, leading to enhanced recognition performance.
However, after reaching this peak, EMG activation typically decreases, particularly in cases where participants are not exerting sustained force to maintain a gesture.
This decline in signal intensity reduces the discriminability of gesture-related features, leading to a degradation in classification performance.
These demonstrate that gesture classification models may benefit from prioritizing peak EMG activity windows rather than relying on prolonged signal durations, which may introduce noise and diminish classification performance.

\subsubsection{Interpretation of the initial peak in opponent gesture prediction}
As shown in~\prettyref{fig:reverse-label}, an early peak is observed when predicting the opponent's actions.
In many trials, one participant initiated movement slightly ahead of the other. One of the players could see the opponent's gesture before completing their own gesture, allowing for time to correct their own gesture to win the match. This might be an explanation for the observable rise in early classification accuracy.
This suggests that gesture prediction in interactive settings is influenced not only by the participant's own motor signals but also by their perception of their opponent's behavior.





\section{Limitations and Future Work}
We used instructed gesture displays to ensure a balanced dataset for training, but the spontaneous gestures in the free-play task introduced greater variability, leading to decreased performance.
Prior studies from the facial expression and speech domains have shown that spontaneous behaviors differ markedly from posed ones in terms of temporal dynamics and data distribution~\cite{tian_emotion_2015, perusquia-hernandez_human_2019, yang_realsmilenet_2020}.
The reduced performance observed in recognizing spontaneous gestures suggests low generalizability to less controlled, real-world inputs.
This highlights the trade-off between ecological validity and data consistency, which poses challenges for achieving reliable recognition.
Another factor that may have contributed to reduced accuracy is variability in sensor alignment and arm pose.
Many participants exhibited differences in arm positioning between tasks, leading to inconsistencies in EMG signal recording.
Even small variations in electrode placement can significantly impact signal quality, affecting the model’s ability to recognize gestures accurately.
Future work could explore adaptive sensor placement techniques or personalized calibration methods to mitigate this issue.
To alleviate muscle fatigue problems in different tasks, future studies should examine the role of fatigue by counterbalancing task order or implementing shorter, interleaved task blocks to better assess its impact on EMG signal stability.
While we analyzed the temporal differences between EMG activation and visible movement, the current method does not account for potential false starts, where participants may initiate a movement before the trial begins but quickly correct and complete the gesture.

\section{Ethical impact statement}
This research was approved by our local committee for research with human data. All participant data was anonymized. All labels are behavioral and annotated by the authors. Reverse label analysis results are contingent on this situation and the two players interacting face-to-face. Generalizations to people other than the dyads playing the game should be investigated in future work.

\section*{Acknowledgment}
M.PH. was supported by the Kakenhi Grants 25K21250 and 22K21309. X.W. was supported by JSPS DC2 Grant 25KJ182900.

\bibliographystyle{IEEEtran}
\bibliography{reference.bib}


\end{document}